\documentclass[letterpaper]{article} 
\usepackage[preprint]{aaai2027}  
\usepackage[hyphens]{url}  
\usepackage{graphicx} 
\usepackage{natbib}  
\usepackage{caption} 
\usepackage{amsmath,amssymb,amsfonts}
\usepackage{booktabs}
\usepackage{multirow}
\usepackage{makecell}
\usepackage{xcolor}
\usepackage{pifont}
\usepackage{xspace}

\definecolor{darkgreen}{RGB}{0,120,60}
\definecolor{darkred}{RGB}{170,40,40}
\newcommand{\cmark}{\textcolor{darkgreen}{\ding{51}}}
\newcommand{\xmark}{\textcolor{darkred}{\ding{55}}}
\newcommand{\cxmark}{\textcolor{orange!85!black}{\ding{52}\rotatebox[origin=c]{-9.2}{\kern-0.7em\ding{55}}}}
\newcommand{\etc}{\emph{etc.}\xspace}

\title{Screenshots or Tools? Eliciting Tool Use and Managing\\
Multimodal Context in Hybrid GUI--MCP Computer-Use Agents}

\author{
    Siqi Fan\textsuperscript{\rm 1},
    Minghao Li\textsuperscript{\rm 2},
    Xiaoqian Ma\textsuperscript{\rm 2},
    Wenhui Tan\textsuperscript{\rm 3},
    Xiusheng Huang\textsuperscript{\rm 2},\\
    Juntong Wu\textsuperscript{\rm 4},
    Liujie Zhang\textsuperscript{\rm 2},
    Shuo Shang\textsuperscript{\rm 1},
    Weihang Chen\textsuperscript{\rm 2}
}
\affiliations{
    \textsuperscript{\rm 1}University of Electronic Science and Technology of China\\
    \textsuperscript{\rm 2}AI Platform, Xiaohongshu Inc.\\
    \textsuperscript{\rm 3}Gaoling School of Artificial Intelligence, Renmin University of China\\
    \textsuperscript{\rm 4}School of Electronic and Computer Engineering, Peking University
}

\begin{document}
\maketitle

\begin{abstract}
Hybrid computer-use agents can act through screenshots or call text tools. We find that
having a tool available does not settle which way the effect goes. Under one identical
GUI--MCP harness on the OSWorld-MCP benchmark (309 tasks), the same MCP tools improve
a reasoning model by $+4.0$pp and degrade a non-reasoning model by $-5.9$pp
(5 runs each, both beyond $2\,$SE). What separates the two is tool-decision behavior. The
non-reasoning policy ignores, misnames, or falsely terminates around tools. The reasoning
model avoids these failures, yet still calls a tool on only 55/309 tasks,
23.9\% of the tool-reachable ones. We call this shortfall the \emph{adoption gap}. Both levels of the
problem share one cause: the model already has a cheaper route and is never trained to
take it. Multi-turn RL probes that cause. At the \textbf{action level}, a dense tool bonus
raises spreadsheet adoption 0.03\,$\to$\,0.33 and carries into greedy decoding,
but held-out accuracy does not follow. Behavior is steerable; competence is not. The
bottleneck lies in tool-call \emph{semantics}. At the \textbf{context level}, a successful tool call often
makes the next screenshot redundant. Dropping it and halving image history cuts input
tokens by about a third, at a small accuracy cost. Retraining under the same observation
rule removes that cost. The compressed agent then reaches 37.8\% against
33.0\% for the uncompressed operating point, at 53\% of the input cost, and
closes the rich--lean gap on a pre-registered degraded subset to zero. Tools help when the model chooses and integrates them, and current hybrid
agents leave many such choices unused.  Code and checkpoints: \url{https://github.com/redai-infra/hybrid-routing-agent}.
\end{abstract}

\section{Introduction}
\label{sec:intro}

Computer-use agents (CUAs) \citep{anthropic2024computeruse,openai2025cua} have two ways to
act on software. They can drive the GUI through screenshots, clicking and typing at image
coordinates. This route is general, but costly and brittle. Each frame consumes vision
tokens, visual history grows over turns, and coordinates expire when the interface changes.
They can also call text-level \emph{tools}: MCP servers \citep{anthropic2024mcp}, CLIs, and
agent ``skills'' \citep{osworldmcp2025,mcpworld2025}. Tools are cheap and precise, but they
exist only for some applications and give no visual confirmation on their own. The choice
between the two routes also sets the serving cost of a deployed agent. Screenshots dominate
the token budget, so every frame kept or dropped is an economic decision as well as a
behavioral one.

Hybrid agents expose both routes. The usual question is whether the tool set or the
injection harness is good enough. We ask a prior question. \emph{When a useful tool is
present, does the model decide to use it?} Our answer is that availability alone does not
fix the sign of the effect. Under one identical retrieval-injection harness on the
OSWorld-MCP benchmark \citep{osworldmcp2025} (309 tasks), the same MCP tools help a
reasoning (``Thinking'') model by $+4.0$pp and hurt a non-reasoning (``Instruct'')
model by $-5.9$pp (Section~\ref{sec:reversal}).

\begin{figure*}[t]
\centering
\includegraphics[width=0.95\textwidth]{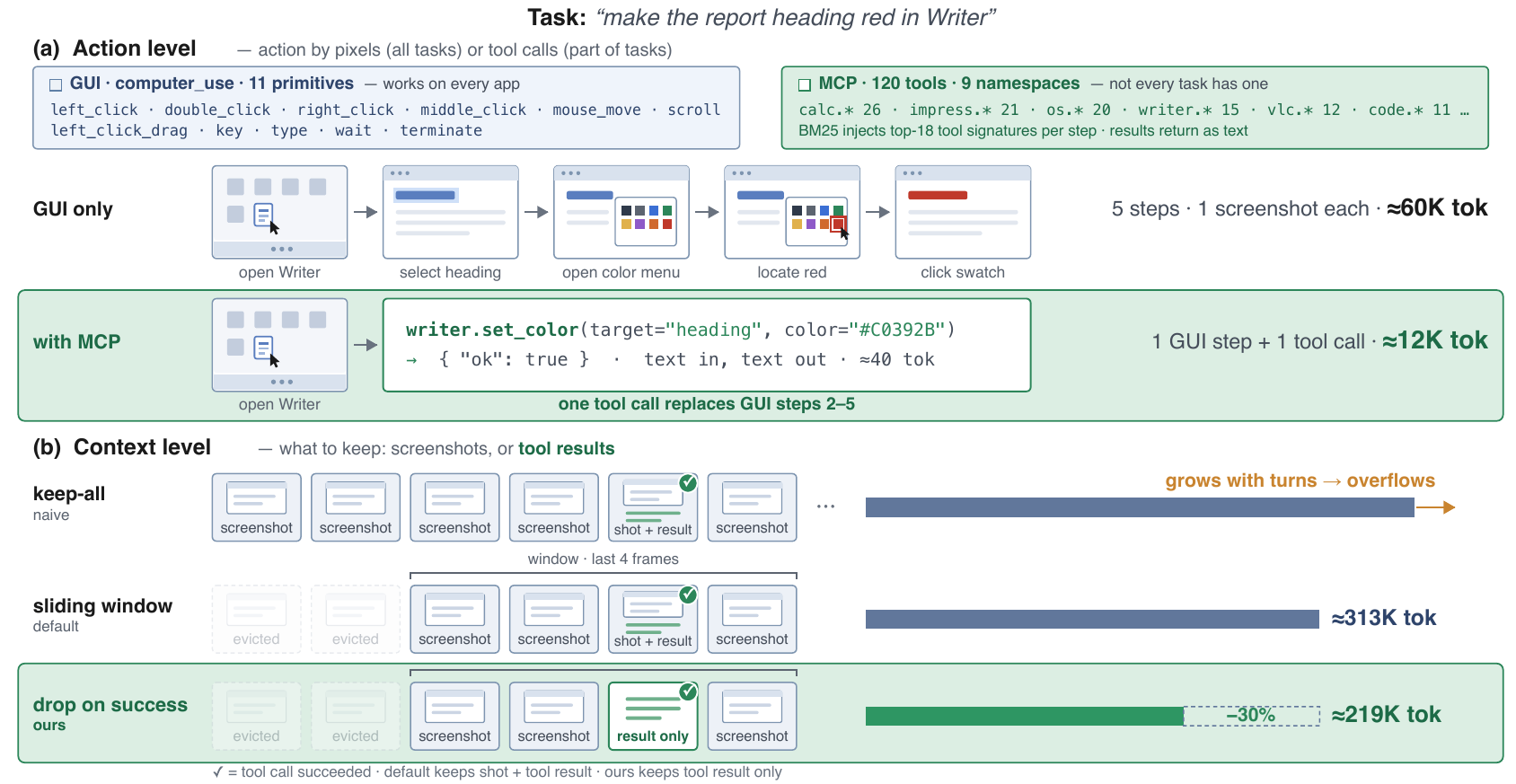}
\caption{\emph{Screenshots or tools?}, asked at two levels for one example task (``make the
report heading red in Writer''). \textbf{(a)} \emph{Action level}: act through pixels or
call a text tool; here a single call replaces GUI steps 2--5.
\textbf{(b)} \emph{Context level}: after a successful tool call, whose result is already in
context as text, choose whether to retain the following screenshot. Token counts are
illustrative.}
\label{fig:overview}
\end{figure*}

This qualifies the degradation reported by concurrent GUI-tool work \citep{toolcua2026}. In
our setting, injection is neither universally harmful nor universally helpful. Its sign
tracks how the model handles the tool decision. The non-reasoning model often ignores,
misnames, or falsely terminates around tools. The reasoning model avoids those failures,
but still uses tools sparingly. Figure~\ref{fig:overview} gives the high-level picture.

\paragraph{The adoption gap.}
The reversal is only the surface. Even the model that benefits \emph{under-adopts}. It
calls a tool on fewer than one task in five. On VLC, where 16/17 tasks are
tool-reachable, it never calls one at all (Section~\ref{sec:adoptiongap}). The model often
\emph{can} use tools but routinely does not, and this shortfall is the central object of
the paper. For practitioners it is a quiet failure mode: the cost of building and injecting
a tool server is paid in full, while most of its benefit goes unrealized.

\paragraph{Screenshots or tools, at two levels.}
We use one recurring question to study this gap: \emph{screenshots or tools?} It appears
twice. At the \textbf{action level}, the agent chooses whether to click through pixels or
call a text tool (Sections~\ref{sec:reversal}--\ref{sec:adoptiongap}; probed with RL in
Section~\ref{sec:rladopt}). At the \textbf{context level}, once a tool has succeeded, the
agent chooses whether to keep the following screenshot or rely on the textual result
(Sections~\ref{sec:frontier}--\ref{sec:context}; probed in Section~\ref{sec:rlctx}).

The two levels are more than an analogy. In both, a cheaper route is already available and
the policy fails to take it, because nothing in training ever asked it to: the tool is
present but unused, the compressed observation affordable but unfamiliar. Vision cannot
disappear either way, since about a quarter of the tasks are tool-unreachable and require
pixels (CAPTCHAs, slide recoloring, heavy in-browser interaction)
\citep{osworldmcp2025,epochai}. The question is therefore not whether to replace screenshots
with tools, but when to use each. The two probes answer differently. The tool decision is
easy to steer, but steering it adds no accuracy. Matching the observation rule between
training and inference does pay off, and makes the compressed configuration the better
deployment point at half the input cost. The next gains in hybrid agents therefore lie less
in adding tools or reward terms than in training signals that teach tool semantics and in
observation rules that match deployment.

\paragraph{Contributions.}
We report a finding, a mechanism for it, and two RL probes that test the mechanism at each
level.
\begin{itemize}
  \item \textbf{Diagnosis: conditional tool effectiveness and the adoption gap.} With the
  same tools and harness, MCP injection helps the reasoning model and hurts the
  non-reasoning one, and the sign tracks tool-decision behavior. Even the model that
  benefits invokes a tool on fewer than a quarter of tool-reachable tasks
  (Sections~\ref{sec:reversal}--\ref{sec:adoptiongap}).
  \item \textbf{Action-level probe.} A dense post-normalization bonus lifts adoption by an
  order of magnitude, and the shift survives into greedy decoding. Held-out accuracy does
  not follow, and a broad sweep does not change that. RL reaches the tool \emph{decision}
  but not tool \emph{competence} (Section~\ref{sec:rladopt}).
  \item \textbf{Context-level probe.} After a successful tool call the next screenshot is
  often redundant. Retraining under the deployment-time observation rule converts the
  compression discount into a half-cost operating point with no out-of-distribution
  accuracy loss (Sections~\ref{sec:context} and~\ref{sec:rlctx}).
\end{itemize}

\section{Related Work}
\label{sec:related}

\paragraph{Computer-use agents \& benchmarks.}
OSWorld \citep{xie2024osworld} provides an execution-based desktop testbed.
OSWorld-MCP~\citeyearpar{osworldmcp2025} adds verified MCP tools and reports that even
strong models invoke tools on only $36.3\%$ of tasks. That low rate is our starting point:
we ask why visible tools go unused, and when using them helps. MCPWorld
\citep{mcpworld2025} studies API/GUI/hybrid evaluation, and OSWorld-Human
\citep{abhyankar2025osworldhuman} temporal efficiency. Mind2Web \citep{deng2023mind2web},
WebArena \citep{zhou2024webarena}, and AndroidWorld \citep{rawles2025androidworld} pose the
same control problem over web and mobile interfaces. OpenCUA \citep{wang2025opencua},
ScaleCUA \citep{liu2025scalecua}, UI-TARS \citep{uitars2025}, and Agent S2 \citep{agents2}
scale CUA data, models, and GUI planning systems \citep{wang2024guisurvey}. We use these
settings to study behavior and token cost.

\begin{table}[!t]
\centering
{\small
\setlength{\tabcolsep}{1.3pt}
\begin{tabular}{@{}lccll@{}}
\toprule
Work & Route & RL & \makecell[l]{Action:\\tool decision} & \makecell[l]{Context:\\management} \\
\midrule
OSWorld-MCP~\citeyearpar{osworldmcp2025} & GUI+MCP & \xmark & \cmark~measured & \xmark \\
ToolCUA~\citeyearpar{toolcua2026} & GUI+tool & \cmark & \cmark~learned & \cxmark~window \\
UltraCUA~\citeyearpar{ultracua2025} & GUI+prog. & \cmark & \cmark~learned & \cxmark~window \\
ComputerRL~\citeyearpar{computerrl2025} & GUI+API & \cmark & \cxmark~implicit & \cxmark~window \\
\midrule
ACON~\citeyearpar{kang2025acon} & text & \xmark & \xmark & \cmark~compress \\
CAT~\citeyearpar{cat2025} & tool calls & \xmark & \xmark & \cmark~callable \\
Context-Folding~\citeyearpar{contextfolding} & tool calls & \cmark & \xmark & \cmark~folding \\
\midrule
\textbf{This work} & GUI+MCP & \cmark & \makecell[l]{\cmark~diagnosed\\\phantom{\cmark}~+ probe} & \makecell[l]{\cmark~matched\\\phantom{\cmark}~rule} \\
\bottomrule
\end{tabular}}
\caption{Positioning against the closest work, along the two levels of the
screenshots-or-tools question. The cross-model \emph{sign} of tool injection and
train--inference context matching are, to our knowledge, examined only here. \xmark: not
addressed; \cxmark: passive or implicit; \cmark: explicitly studied or learned.}
\label{tab:positioning}
\end{table}

\paragraph{GUI-tool hybrid agents \& modality.}
ToolCUA \citep{toolcua2026} is the most directly concurrent work. It trains GUI-tool path
orchestration with heavy RFT/RL ($46.85\%$ on OSWorld-MCP) and reports that naive MCP
injection can hurt a GUI agent. Our result narrows that observation. The sign of injection
tracks the base model's tool-decision behavior, which our RL experiment then isolates.
UltraCUA \citep{ultracua2025} likewise trains hybrid
GUI/programmatic actions at scale, and the tool route itself builds on LLM tool learning
\citep{schick2023toolformer,qin2024toolllm,patil2024gorilla}. A separate line improves the
screenshot side of the interface, covering grounding, GUI perception, and action modeling:
CogAgent \citep{hong2024cogagent} UGround
\citep{gou2024uground}, Aguvis \citep{xu2025aguvis}, MP-GUI \citep{mpgui}, and OS-Atlas
\citep{osatlas2024}.

\paragraph{RL for agents \& credit assignment.}
ComputerRL \citep{computerrl2025} scales online RL to $48.9\%$ on OSWorld with a single
large run over an API--GUI hybrid action space. DigiRL \citep{bai2024digirl} and WebRL
\citep{qi2025webrl} train device-control and web agents with online RL. GUI-R1, UI-R1, and
UI-TARS-2 \citep{xia2025guir1,lu2025uir1,uitars22025} apply RL to GUI action prediction and
multi-turn control. Our probe targets a different question: whether an outcome-level signal
can make the policy leave the GUI path for a tool it already has. Our advantage follows the group-relative (R1-zero) line
\citep{deepseekr1,shao2024deepseekmath,liu2025drgrpo,dapo2025}, a critic-free
simplification of PPO \citep{schulman2017ppo}, with a per-step broadcast that keeps long
trajectories from dominating the gradient. Turn- and segment-level credit assignment
\citep{mtgrpo,spo,turnppo,agentwisenorm,areal} remains complementary.

\paragraph{Context management for long-horizon agents.}
Long-horizon agents manage bounded context with memory hierarchies
\citep{packer2023memgpt} and prompt compression \citep{jiang2023llmlingua}. ACON
\citep{kang2025acon} compresses single-modality \emph{text} context. We study the same
pressure in a dual-modality setting, where old information can live as pixels or as tool
text. CAT \citep{cat2025} and Context-Folding \citep{contextfolding} make context
maintenance a callable or learned operation. We instead keep the rule fixed, which isolates
the train/inference mismatch before any context policy is learned.
Table~\ref{tab:positioning} summarizes the closest comparisons.

\section{Action Level: Does the Model Use Its Tools?}
\label{sec:action}

\paragraph{Setup.}
We compare two checkpoints of the same 8B backbone, Qwen3-VL-8B-Thinking (reasoning) and
Qwen3-VL-8B-Instruct (non-reasoning) \citep{qwen3vl}, which differ in whether they emit an
explicit reasoning trace. The benchmark is \texttt{test\_all\_no\_internet} (309 tasks) from
OSWorld-MCP~\citeyearpar{osworldmcp2025}, built on OSWorld \citep{xie2024osworld}.
Everything else is held fixed: harness, retriever, prompt template, and tool set. The MCP
inventory spans 120 tools in 9 application namespaces, exposed through BM25
\citep{robertson2009bm25} top-18 retrieval with one call per step and structured error
feedback. Retrieval is keyed to the active application, so multi-app tasks see a
toolset that changes as the agent switches apps. Qwen3-VL emits relative coordinates on a
1000-grid, and the harness applies the required resize (Appendix~A). All evaluations use greedy decoding, \texttt{max\_steps}$=50$, and
five repeated runs. We call a difference significant only when $|\Delta|>2\,\mathrm{SE}$.

\paragraph{Unified hybrid action space.}
GUI actions and MCP tools are presented through the same call surface. At each step the model
emits one \texttt{<tool\_call>} object: either \texttt{computer\_use} with one of its
11 primitive actions (click, double-click, drag, scroll, key, type, wait, terminate,
\etc), or one of the retrieved MCP tools. Both appear in the same
\texttt{<tools>} block. No external controller routes the model. It must choose inside one
action head whether to act through pixels or through a text tool
\citep{yao2023react,schick2023toolformer}.

\paragraph{Context construction.}
A pure-GUI agent keeps its full textual action history plus a sliding window over the last
$k$ screenshots, the standard bounded-memory recipe for CUAs
\citep{xie2024osworld,uitars2025}. Adding MCP tools leaves this skeleton unchanged but opens
a second, textual observation channel. Formally, the step-$i$ assistant output
$a_i=(\texttt{think}_i,\,\text{Action}_i,\,c_i)$ consists of an optional reasoning trace
(Thinking model only), a one-line action summary, and exactly one call $c_i$. Each user turn
is $u_i=\rho(r_i)\oplus \tilde o_i$, where $r_i$ is the (possibly empty) result of $c_{i-1}$
truncated to $L{=}1500$ characters, and $\tilde o_i$ is the retained screenshot. The context
at step $t$ is
\begin{equation}
\begin{split}
C_t \;=\; \sigma(\mathcal{T}_t)
\;\oplus\; &\underbrace{\bigoplus_{i=\tau_t}^{t-1}\!\bigl(u_i \oplus a_i\bigr)}_{\text{windowed visual memory}} \\[2pt]
\;\oplus\; &\underbrace{u_t \oplus I \oplus H_t}_{\text{current turn}},
\qquad \tau_t=\max(1,\,t{-}k{+}1).
\end{split}
\label{eq:context}
\end{equation}
Here $\sigma(\mathcal{T}_t)$ is the system message, holding the format rules and the
BM25-retrieved tool set $\mathcal{T}_t$, and $I$ is the task instruction. The text trace
$H_t=\bigoplus_{i<t}\text{Action}_i$ keeps the one-line summary of every prior step
\citep{yao2023react,shinn2023reflexion}. A raw screenshot $o_i$ costs roughly 2K vision
tokens, and $\tilde o_i$ is the version kept after the retention rule in Eq.~\ref{eq:skip}
(rendered prompt in Appendix~C). The two memories run at different timescales.
Pixels are windowed at depth $k$ (default $k{=}4$) while text persists in full, so old
pixels can drop out while their semantic trace and tool results $r_i$ remain. This asymmetry
motivates the two knobs of Section~\ref{sec:ctxcost}.

\begin{table*}[t]
\centering
{\small
\setlength{\tabcolsep}{2.5pt}
\begin{tabular}{@{}lcccccccccccc@{}}
\toprule
& \multicolumn{5}{c}{Thinking} & \multicolumn{5}{c}{Instruct} & \multicolumn{2}{c}{Adoption$^{\dagger}$} \\
\cmidrule(lr){2-6}\cmidrule(lr){7-11}\cmidrule(lr){12-13}
& GUI & \multicolumn{4}{c}{GUI{+}MCP, by context policy} & GUI & \multicolumn{4}{c}{GUI{+}MCP, by context policy} & Think & Inst \\
\cmidrule(lr){3-6}\cmidrule(lr){8-11}
Domain & win.~4 & \makecell{win.~4\\(op.)} & \makecell{win.~4\\$+$drop} & win.~2 & \makecell{win.~2$+$drop\\(ctx\_opt)} & win.~4 & win.~4 & \makecell{win.~4\\$+$drop} & win.~2 & \makecell{win.~2\\$+$drop} & & \\
\midrule
calc (spreadsheet) & 17.9 & 18.3 & 14.9 & 12.8 & 12.3 & 17.0 & 7.2 & 5.5 & 3.8 & 4.7 & 21\% & 0\% \\
writer (document)  & 51.3 & 45.2 & 38.3 & 37.4 & 38.3 & 29.9 & 14.8 & 13.9 & 5.2 & 9.6 & 52\% & 22\% \\
impress (slides)   & 28.5 & 26.8 & 20.9 & 21.7 & 20.0 & 18.5 & 15.3 & 17.9 & 16.6 & 10.6 & 40\% & 40\% \\
vs\_code           & 57.1 & 68.6 & 64.8 & 62.9 & 64.8 & 49.5 & 38.1 & 39.0 & 43.8 & 45.7 & 14\% & 19\% \\
os                 & 51.7 & 57.5 & 57.5 & 53.3 & 54.2 & 41.7 & 45.0 & 43.3 & 46.7 & 47.5 & 8\% & 4\% \\
multi\_apps        & 8.3 & 14.9 & 14.1 & 13.6 & 13.6 & 7.0 & 9.9 & 10.9 & 9.6 & 9.1 & 11\% & 4\% \\
\midrule
\multicolumn{13}{@{}l}{\emph{Zero-adoption domains (accuracy changes not tool-attributable):}}\\
gimp               & 60.8 & 67.7 & 70.0 & 69.2 & 66.9 & 57.7 & 34.6 & 38.5 & 50.8 & 46.9 & 0\% & 0\% \\
thunderbird        & 6.7 & 10.7 & 12.0 & 8.0 & 12.0 & 6.7 & 4.0 & 4.0 & 6.7 & 6.7 & 0\% & 0\% \\
vlc                & 24.7 & 40.0 & 34.1 & 31.8 & 31.8 & 27.1 & 16.5 & 25.9 & 25.9 & 21.2 & 0\% & 0\% \\
chrome             & 64.3 & 68.6 & 74.3 & 64.3 & 64.3 & 62.9 & 54.3 & 61.4 & 44.3 & 44.3 & 0\% & 0\% \\
\midrule
\textbf{all (309)} & \textbf{30.5} & \textbf{34.5} & 32.3 & 30.6 & 30.6 & \textbf{25.4} & \textbf{19.5} & 20.9 & 20.5 & 19.5 & 17.8\% & 10.4\% \\
\midrule
\multicolumn{13}{@{}l}{\emph{Token side (per task, 5-run means):}}\\
input (K)        & 313.1 & 337.1 & 342.4 & 226.1 & \textbf{219.5} & 287.3 & 316.4 & 310.1 & 232.4 & 231.1 & --- & --- \\
peak (p95)       & 11385 & 11544 & 11487 & 7314 & \textbf{7243} & 10966 & 11441 & 11437 & 7254 & 7231 & --- & --- \\
tok / 1\%acc (K) & 10.6 & 10.1 & 11.0 & 7.8 & \textbf{7.7} & 11.4 & 16.3 & 14.9 & 11.5 & 12.0 & --- & --- \\
\bottomrule
\end{tabular}}
\caption{The static results grid: accuracy (\%, per domain and overall) and token cost
across the action space (GUI vs.\ GUI+MCP) and the context policies, named as in
Figure~\ref{fig:overview}b (win.~$k$ $=$ sliding window over the last $k$ screenshots;
$+$drop $=$ drop-on-success, Eq.~\ref{eq:skip}); 5-run means. ``op.''\ marks the RL
operating point and ``ctx\_opt'' the compression setting of Section~\ref{sec:context}.
$^{\dagger}$task-level tool invocation, measured at the operating point. Tok/1\%acc
counts input$+$output.}
\label{tab:perapp}
\end{table*}

\paragraph{Retention rule (the second knob).}
The retained frame $\tilde o_i$ realizes the screenshot-retention decision:
\begin{equation}
\tilde o_i \;=\;
\begin{cases}
\pi & \text{if \textsc{drop} is on and } \mathrm{succ\_mcp}(c_{i-1}),\\[2pt]
o_i & \text{otherwise,}
\end{cases}
\label{eq:skip}
\end{equation}
where $\pi$ is a short text placeholder. Here $\mathrm{succ\_mcp}$ denotes
\emph{execution-level} success: the call parsed, dispatched, and returned without error.
Semantic success is a separate matter, and the distinction returns in
Section~\ref{sec:context}. Image-history depth $k$ and this drop rule are the two context
knobs we vary below. Both act only on the visual channel and leave the text trace $H_t$
intact.

\subsection{The Sign Reversal}
\label{sec:reversal}

\paragraph{Overall result.}
With everything except the base model held fixed, MCP injection lifts the reasoning model
and drops the non-reasoning one (Table~\ref{tab:perapp}, all-309 row; both deltas beyond
$2\,\mathrm{SE}$). The best single Thinking run reached 37.9\%; we report five-run
means throughout.

\paragraph{Per-domain decomposition.}
The same pattern holds per domain (Table~\ref{tab:perapp}). Among the six domains with
nonzero adoption, tools help the reasoning model on four and hurt the non-reasoning model on
four. We exclude the zero-adoption block from this count, since with no tool calls its
changes reflect only prompt perturbation and run-to-run variance. Adoption alone is not
enough, though. The reasoning model's two losses are its highest-adoption domains. Writer is
the sharpest case: half of its tasks invoke a tool, yet invoked tasks succeed far less often
than non-invoked ones, a mix of difficulty self-selection and mis-parameterized calls
(Appendix~E). The non-reasoning model is hurt most precisely where it nominally
adopts.

\begin{table}[!tb]
\centering
{\small
\setlength{\tabcolsep}{3pt}
\begin{tabular}{lcc}
\toprule
Diagnostic & Thinking & Instruct \\
\midrule
\multicolumn{3}{@{}l}{\emph{Supply $\to$ adoption:}}\\
Tool-reachable (supply) & \multicolumn{2}{c}{230/309 (79 vision-only)} \\
Adoption, task-level        & 17.8\% (55/309) & 10.4\% (32/309) \\
Adoption, reachable (230)   & 23.9\% & 13.9\% \\
TIR$_{\text{real}}$ (MCP/steps) & 2.8\% & 2.0\% \\
\midrule
\multicolumn{3}{@{}l}{\emph{Failure modes:}}\\
Hallucinated MCP steps      & $+0.0$pp & $+0.4$pp \\
False-success rate          & 21.7\% (67/309) & 33.0\% (102/309) \\
Hallucinated tool names     & 0 & 97 (2 tasks) \\
\bottomrule
\end{tabular}}
\caption{Behavioral diagnostics at the operating point (GUI+MCP, window-4; 5-run means).
Instruct adopts fewer tools
yet false-terminates \emph{more}, ruling out a ``cleaner SFT prior'' explanation.
TIR $=$ MCP steps / total steps. Token costs: Table~\ref{tab:perapp}, token block.}
\label{tab:diagnostics}
\end{table}

\paragraph{What differs between the two models.}
\label{sec:mechanism}
The reversal tracks how the two models handle tools (Table~\ref{tab:diagnostics}). The
non-reasoning model ignores the spreadsheet tools entirely, hallucinates tool names, and
false-terminates \emph{more} often, all while producing $6$--$7\times$ shorter outputs. The
hallucinations concentrate in two tasks, and fixing them would barely move the total, so
they are a symptom and not the cause. The pattern instead suggests that without an explicit
deliberation trace \citep{wei2022cot}, the model never takes the step of asking whether a
tool should be used. Correctly injected tools are then ignored, misnamed, or hidden behind
premature success. The two checkpoints differ in more than that trace, so we read this as
association and not mechanism.

\subsection{The Adoption Gap}
\label{sec:adoptiongap}

Even the reasoning model leaves most tools unused. It invokes a tool on fewer than one task
in five overall, and on fewer than one in four where a tool is actually reachable
(Table~\ref{tab:diagnostics}). Removing the capability ceiling does not remove the behavior
gap.

The two losses separate cleanly (Table~\ref{tab:diagnostics}, supply block). About a quarter
of the tasks are tool-unreachable: three apps expose no MCP tools at all, and some tasks in
tool-equipped apps have no tool that applies. Everywhere else the tools are present and
injected, yet mostly unused. VLC is the extreme case, where nearly every task is
tool-reachable and neither model ever calls one. Section~\ref{sec:rladopt} targets this
purely behavioral gap.

\section{Context Level: What Does Hybrid Cost?}
\label{sec:ctxcost}

Section~\ref{sec:action} asked whether the agent takes the tool route; this section asks
what each route costs. Screenshots dominate the token budget, so the window depth $k$ and
the drop rule of Eq.~\ref{eq:skip} set the serving price of a hybrid agent.

\subsection{Accuracy--Token Frontier and Operating Points}
\label{sec:frontier}

Image-history depth: \emph{window-4} keeps the last four screenshots, \emph{window-2} the
last two. Post-tool retention: by default the
next screenshot is kept, while \emph{drop} (drop-on-success, Eq.~\ref{eq:skip}) replaces it
with a text placeholder. This gives five operating points: GUI-only; window-4, our
accuracy-oriented hybrid baseline; window-2; window-4$+$drop; and window-2$+$drop, the
token-efficient ``ctx\_opt'' setting.

The two knobs do different jobs (Table~\ref{tab:perapp}, token block). Window depth is the
main token lever. Window-2 cuts cumulative input by about a third and peak context by nearly
40\%, but it also causes the only accuracy loss beyond $2\,\mathrm{SE}$. The drop rule
is nearly free in accuracy at both depths, because it removes a frame whose tool result is
already in text. On its own it is not a token lever, however. At window-4 the slightly
longer completions offset the per-frame saving, so its benefit appears only alongside the
shorter window. Together, window-2$+$drop is the token-efficiency knee.

We keep \textbf{window-4} (no drop) as the RL operating point and \textbf{window-2$+$drop}
as the compression studied in Section~\ref{sec:context}.

\subsection{Inference-Only Compression}
\label{sec:context}

Applied only at inference, the token-efficient setting of Section~\ref{sec:frontier}
(window-2$+$drop, ``ctx\_opt'') costs $-3.9$pp ($\pm1.0$). Paired per-task analysis
points to \emph{diffuse degradation} rather than lost capability. Only 3/309 tasks
flip hard under compression, while a pre-registered \emph{degraded subset} D13 (13 tasks,
with a 12.8pp rich--lean gap under matched greedy anchors) concentrates the effect.
Our hypothesis is therefore mis-adaptation, not incapability, and not simply a
horizon-budget issue \citep{kang2025acon}. The policy is asked to act on an observation
distribution it never saw during training. If that is right, the loss should be recoverable by
making rollouts, evaluation, and deployment share one observation policy, which
Section~\ref{sec:rlctx} tests.

\paragraph{Scope.}
$\mathrm{succ\_mcp}$ is execution-level, so a call can succeed mechanically yet fail
semantically. A find-and-replace may return success with zero replacements, and the drop
rule then discards the only visual evidence of that failure.
Section~\ref{sec:rladopt} shows this is common on parameter-heavy tools. Part of the
residual drop cost is therefore lost error-correction signal, not merely a dropped
duplicate.

\section{Multi-Turn RL: Steering Tool Use and Matching the Observation Rule}
\label{sec:elicit}

RL lets us probe both levels directly. At the action level, can an outcome-level signal make
the policy use tools it already has? At the context level, can matched training recover the
compression penalty? All runs use the verified-clean pipeline of Appendix~A and are on-policy
by construction (per-window \texttt{clip\_frac}$=0$,
ratio $=1.0$).

\subsection{Setup}
\label{sec:rlsetup}

\paragraph{Multi-turn GRPO.}
For each task $x$ we roll out $G{=}8$ trajectories at temperature $1.0$ across 96 parallel
environments. All 74 curated training tasks (below) are rolled out at every step, with
horizon $T_{\max}{=}\texttt{max\_steps}{=}50$. Evaluation uses greedy decoding. A trajectory $\tau$
with $T$ steps and terminal outcome $\mathrm{succ}(\tau)\in\{+1,-1\}$ receives the return
\begin{equation}
R(\tau) \;=\; \mathrm{succ}(\tau)
\;-\; \lambda_{\text{len}}\frac{T}{T_{\max}}
\;-\; \lambda_{\text{cap}}\,\mathbb{1}\!\left[T\ge T_{\max}\right],
\label{eq:return}
\end{equation}
with tie-breaker coefficients $\lambda_{\text{len}}{=}0.05$ and
$\lambda_{\text{cap}}{=}0.2$; the $\pm1$ outcome term dominates. Returns are $z$-scored
within the group ($\mu_x,\sigma_x$) and broadcast uniformly to steps. This is a Dr.GRPO-style
length debiasing \citep{liu2025drgrpo} that keeps long trajectories from dominating the
gradient. Each step then becomes one training sample, with prompt $C_t$
(Eq.~\ref{eq:context}) and response $a_t$:
\begin{equation}
\hat{A}_t \;=\; \frac{1}{T}\cdot\frac{R(\tau)-\mu_x}{\sigma_x}
\;+\; \lambda_{\text{mcp}}\,b_t .
\label{eq:adv}
\end{equation}

\paragraph{A dense tool bonus that survives normalization.}
The bonus $b_t\in\{0,1\}$ (used in Section~\ref{sec:rladopt}) is added \emph{after}
normalization. Placed inside $R(\tau)$ it would be diluted to ${\sim}10^{-4}$ by trajectory
averaging, $z$-scoring, and the $1/T$ broadcast, which is empirically a dead signal. It
fires only for an execution-successful, non-read-only call whose (tool, arguments) key has
not appeared earlier in the trajectory. Firing once per key per trajectory, and not gating
on task success, prevents reward farming by repeated or side-effect-free calls. With
$\lambda_{\text{mcp}}{=}0.1$ the bonus is louder than the primary signal at the steps where
it fires, since the outcome-derived part of $\hat{A}_t$ has magnitude ${\sim}0.07$ after
normalization.

\paragraph{Optimization.}
We use GRPO with a KL penalty to the \emph{rollout-time} policy ($\beta{=}0.02$; hyperparameters in
Appendix~F). Anchoring to the base model
instead cancels the bonus, because the per-step drift it induces is smaller than the pull
back to base. Groups with mean success outside $(0.05,0.95)$ are dropped, so every kept
group mixes successes and failures. Each configuration is trained once; reported
accuracies are greedy probes repeated three times.

\paragraph{Data split and gradient-band curation.}
The eight trained apps (calc, writer, impress, vs\_code, os, gimp, vlc, thunderbird)
contribute a 172-task training pool and 48 held-out tasks; chrome and
multi\_apps (89 tasks) are never trained on, a true out-of-distribution bucket. From this
pool we curate the actual training set: a \emph{gradient band} of 74 tasks with empirical
pass rate $p\in(0.1,0.9)$ under the same $G{=}8$ temperature-1.0 rollouts. Groups outside
the band have zero within-group variance and contribute zero gradient, so pool tasks outside
the band are never rolled out. The band is profiled once before launch and held fixed for
all runs reported here, which leaves 235 tasks never trained on.

\paragraph{Outcome-only RL.}
With $\lambda_{\text{mcp}}{=}0$, no swept configuration moves held-out or out-of-distribution
accuracy. In-distribution accuracy does rise (Section~\ref{sec:rlctx}); nothing transfers.
The sweep covers learning rate, KL strength and anchor, normalization scheme, task density,
and horizon (Appendix~F). The signal is simply too sparse: a long-horizon task
yields a single $\pm1$ over up to 50 steps, and behaviors the base policy never samples,
above all successful calls to the harder tools, receive no gradient under R1-zero-style RL.
Any movement reported below therefore comes from the dense bonus
(Section~\ref{sec:rladopt}) or from training on the gradient band itself. A matched
\emph{rich-observation} control (identical recipe, compression off, evaluated at the same
30-step checkpoint) settles the attribution in Section~\ref{sec:rlctx}.

\subsection{Result 1: A Dense Tool Bonus Moves Adoption, Not Competence}
\label{sec:rladopt}

\paragraph{The tool decision is fully steerable.}
With the dense bonus, applied on a 24-task subset of the gradient band, spreadsheet adoption
rises from 0.03 to 0.33 within 23 training steps. It transfers to greedy
decoding (0.02\,$\to$\,0.29), so the change is learned policy and not sampling
noise, and step-level usage rises 4.7$\times$. One post-normalization reward term is
enough to change the decision that Section~\ref{sec:adoptiongap} showed models rarely make.

\begin{figure}[!tb]
\centering
\includegraphics[width=\columnwidth]{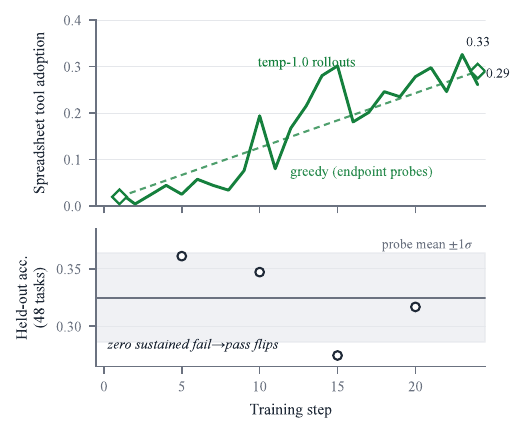}
\caption{Adoption--competence decoupling: the dense tool bonus drives adoption into the
deterministic policy (top) while held-out accuracy stays at the base level (bottom);
dashed connector: visual guide, not per-step data.}
\label{fig:decouple}
\end{figure}

\paragraph{Held-out accuracy localizes the bottleneck.}
Across the 48 held-out tasks the run produces zero sustained fail$\to$pass flips, and on
the seven held-out spreadsheet tool tasks accuracy stays at the base level throughout
(Figure~\ref{fig:decouple}). The probe therefore separates the tool \emph{decision},
which RL controls, from tool-call \emph{competence}, which it leaves untouched.

\paragraph{Diagnosis: calls execute but fail semantically.}
Tool calls execute reliably (98--100\% API success), but on parameter-heavy tools the
\emph{semantic} success rate is zero: 0/23 for regex find-and-replace and 0/16
for format conversion. The server reports \texttt{success:true} on zero-effect calls, such
as a regex that matches nothing, which feeds a false-success prior. This is the same
execution-versus-semantics gap that limits the drop rule (Section~\ref{sec:context}).
Within-task matched comparisons show no win-rate difference with tools ($-3.1$pp, n.s.).
Three mechanisms give the same result: an outcome-independent bonus (RL),
positive-advantage cloning (RL), and tool documentation injected at inference time
(prompting) all raise adoption substantially (hint-targeted calls rise $3\times$), and none
changes accuracy. The bottleneck is tool-call semantics,
which makes it a data problem and not a reward-design problem.

\subsection{Result 2: Keeping the Token Savings Without the Accuracy Loss}
\label{sec:rlctx}

\begin{figure}[t]
\centering
\includegraphics[width=0.93\columnwidth]{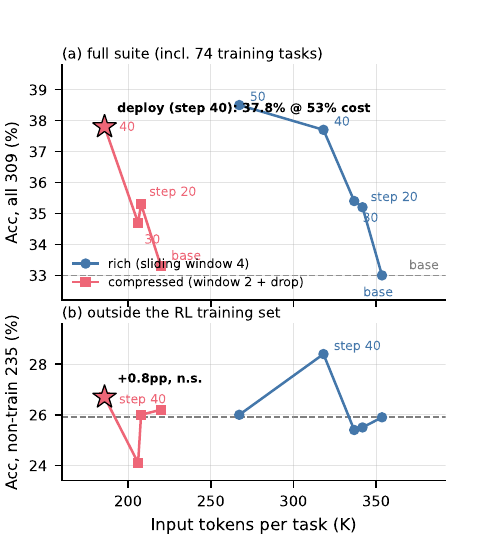}
\caption{Final deployment plane: RL checkpoints under both observation policies
(greedy$\times$3 anchors; point labels are training steps; star $=$ the step-40
deployment pick) on \textbf{(a)} the full suite and \textbf{(b)} the 235 tasks
outside the RL training set. The step-50 compressed point is omitted (token telemetry
unavailable).}
\label{fig:deploy}
\end{figure}

\begin{figure}[t]
\centering
\includegraphics[width=\columnwidth]{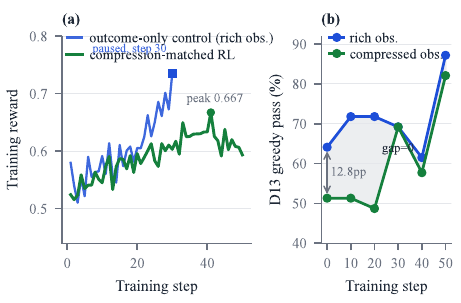}
\caption{Matched-training recovery. \textbf{(a)} Training reward against the
rich-observation control (paused at step 30 after probing).
\textbf{(b)} D13 under both observation policies (shaded: the rich--lean gap).}
\label{fig:recovery}
\end{figure}

\paragraph{Design.}
We reuse the recipe of Section~\ref{sec:rlsetup} with $\lambda_{\text{mcp}}{=}0$. The
\emph{only} change is the observation policy. The compression rule of
Section~\ref{sec:context} is now active in both rollout and evaluation, logged and replayed
so that train and inference share it exactly. Two judgment criteria were pre-registered
before launch: the degraded subset D13 (the 13 tasks where inference-only compression
concentrates its loss), and a difference-in-differences (DiD) criterion requiring the
compressed-side gain to exceed the rich-side gain by $\geq$15pp, which subtracts
memorization common to both sides.

\paragraph{Result.}
The training reward rises 0.52\,$\to$\,0.667 (peak, step 41). What matters is
not the rise itself, since in-distribution gains can contain memorization, but what it is
made of. Checkpoints at steps 20--50 are probed on all 309 tasks under \emph{both}
observation policies against same-period base anchors (greedy$\times$3, each within the
operating point's 5-run spread):
\begin{itemize}
\item \textbf{Deployment headline (Figure~\ref{fig:deploy}).}
Matched training removes the price of compression. The step-40 compressed checkpoint runs at
53\% of the input cost, with $-37$\% peak context, and loses nothing outside the
training set ($+0.8$pp, n.s., on the 235 non-training tasks): equal accuracy at
half the cost. On the full suite it reaches 37.8\% against 33.0\% for the
uncompressed base operating point, $+4.8$pp (step 50 holds $+3.9$pp), but
${\sim}4.1$pp of that margin comes
from the 74 training tasks (55.8\%$\,\to\,$72.9\%), so we read the full-suite
number as in-distribution. The optimum is also an \emph{early} checkpoint, not the endpoint.
\item \textbf{Degraded subset D13.} The rich--lean gap collapses to 0 at step 30 and
stays closed within noise ($3.8$--$5.1$pp; Figure~\ref{fig:recovery}b), with both sides
ending far above base (compressed 51.3\%$\,\to\,$82.1\%, rich
64.1\%$\,\to\,$87.2\%). Of the two pre-registered criteria, the
\emph{direction} holds and the \emph{magnitude} does not. The DiD turns positive at step 30
($+12.8$pp) and stays positive thereafter ($+7.7$--$+9.0$pp, $\pm{\sim}11$pp,
$n{=}13$), but the $+15$pp bar is met at no checkpoint. That bar was calibrated against
single-rep anchors which, we now know, overstated the base gap; we record it as failed
rather than restate it at a revised threshold.
\item \textbf{Late training favors the rich side.} After the compressed side adapts at
step 30, further training helps mainly under rich observation (on training tasks,
$+10.9$pp from step 30 to 50 versus flat compressed), hence the \emph{early}
step-40 optimum.
\end{itemize}

\paragraph{Scope and attribution.}
The gains above are in-distribution. Held-out accuracy under compression is unchanged
between the base and the final checkpoint. The rich-observation control settles where the
training-side rise comes from. At the same 30-step checkpoint it gains nearly twice as much
on the training band ($+20.5$pp vs.\ $+11.5$pp; on D13, $+12.8$pp vs.\ $+5.1$pp). The rise
is therefore a property of the optimization recipe and not of consistency training, and
compression is if anything a training-time handicap. Both the deployment result and the D13 dynamics
are measured within the compressed run, so neither depends on the control. The control's
faster learning is also
not free: it reaches 41.0\% at step 30, but emits 42.6K output tokens per task
against the compressed run's 10.7K and keeps the full uncompressed input cost, about
$1.6\times$ the compressed total. The two runs therefore sit at non-comparable points of the
cost--accuracy frontier. Near-lossless is also not lossless: spreadsheet and slide tasks
can still need visual confirmation.

\section{Conclusion}
\label{sec:conclusion}

Hybrid GUI--MCP agents face the same question twice: \emph{screenshots or tools?} At the
action level, tools help only when the model decides to use and integrate them. The same
injection helps the reasoning model and hurts the non-reasoning one, RL steers the decision
freely, and tool-call semantics remains the open problem. At the context level, dropping
redundant post-tool screenshots pays off only when training and inference share one
observation rule; matched retraining then halves input cost with no out-of-distribution
accuracy loss. Both levels have one shape: a cheaper route exists and nothing in training
teaches the policy to take it. Tool availability is not enough, and current hybrid agents
leave many usable tools on the table.

\paragraph{Limitations and future work.}
The sign reversal is bounded: within one backbone, the sign of MCP injection tracks
tool-decision behavior, but we do not show that reasoning capability sets that sign in
general. The two checkpoints differ in more than their reasoning trace, and a within-model
thinking toggle would settle the mechanism. The RL probe
identifies the competence bottleneck but does not close it. Supervised injection of verified tool
trajectories from stronger teachers is the next step, with turn- and segment-level credit
assignment \citep{mtgrpo,spo,turnppo} complementary. The
compression-recovery gain does not transfer to held-out tasks, and the in-distribution rise
is a recipe effect (Section~\ref{sec:rlctx}). The context rule is fixed rather than learned
\citep{cat2025}, and cross-OS scaling \citep{liu2025scalecua} is left to future work.

\bibliography{references}

\clearpage
\appendix

\section{Harness Correctness}
\label{sec:appendix-harness}

The reasoning-gating claim presumes a \emph{correct} harness, which we separate from the
tool-surfacing design that the claim is about. Correctness has three requirements. (1) The
action parse format must match the model's native \texttt{computer\_use} calling pattern. (2)
The coordinate convention must match the model: Qwen3-VL emits \emph{relative} coordinates
on a $1000$-grid and requires a resize that absolute-pixel models do not. (3) The VM--agent
feedback loop (action execution $\to$ observation/error return) must be closed and
deterministic. If any of these is wrong, observed failures stem from format rather than
capability and the GUI-vs-MCP comparison is unfair.

\paragraph{Verification.}
We check these requirements directly rather than by inspection. Training is on-policy by
construction: per-window \texttt{clip\_frac}$=0$ and ratio $=1.0$. A saved checkpoint is
byte-identical to the base model except for the weights, so no configuration drift enters
across resumes. Every RL run reported in the paper uses this verified pipeline.

\section{Context-Construction Strategy Menu}
\label{sec:appendix-context}

Table~\ref{tab:ctxmenu} lists the full set of context-construction strategies as choices over
image-history depth $k$ and the screenshot retention rule in Eq.~(1) of the main paper. The first
three rows correspond to the five harness configurations evaluated in Section~4.1 of the main paper;
the lower rows are natural extensions along the same axes.

\begin{table*}[tp]
\centering
\small
\begin{tabular}{p{3.2cm}p{5.0cm}p{4.0cm}}
\toprule
Strategy & Construction & Cost / trade-off \\
\midrule
full & all screenshots $+$ all responses & maximal info; context explodes \\
baseline (default) & window of last $k{=}4$ screenshots; older steps as text & standard bounded memory \\
skip\_on\_mcp\_success & within the window, replace the screenshot after a successful MCP call with a text placeholder & near-lossless; trims costliest frames \\
\midrule
\multicolumn{3}{l}{\emph{Extensions along the same axes:}}\\
window $k{=}1$ & keep only the current screenshot & cheapest; loses spatial memory \\
state-change skip & drop a frame byte-identical to the previous one & more general than MCP-gated skip \\
downscale / a11y tree & shrink old frames or swap for accessibility text & further token cuts \\
\bottomrule
\end{tabular}
\caption{Context-construction strategy menu. Image-history depth $k$ and retention rule
(whether to keep or replace the screenshot with a placeholder) are the two axes.
The first three strategies correspond to the five configurations in Section~4.1 of the main paper.}
\label{tab:ctxmenu}
\end{table*}

\section{Concrete Context Template}
\label{sec:appendix-template}

The message array instantiating Eq.~(1) of the main paper interleaves one \texttt{system} message
with alternating \texttt{user}/\texttt{assistant} turns; within the depth-$k$ window each
\texttt{user} turn carries an optional relayed tool result $\rho(r_i)$ and the retained frame
$\tilde o_i$, and the \emph{current} turn additionally appends the instruction $I$ and the full
action trace $H_t$. GUI and MCP calls share one \texttt{<tool\_call>} schema
(the ``unified action space'').

\begin{footnotesize}
\begin{verbatim}
system   : # Tools <tools>{computer_use}
         {libreoffice_calc.set_cell_value}
         ...BM25 top-18 signatures...
         </tools> + format rules
user     : [Tool result: ...]?
         + screenshot_i  <- windowed step i
assistant: (<think>...</think>)? Action: ...
         <tool_call>{...}</tool_call>
\end{verbatim}
\begin{verbatim}
...        (last k steps kept)
user     : [Tool result: ...]?
         + current screenshot
         + Instruction: <I>
         + Previous actions: Step 1..t-1
           (H_t)
\end{verbatim}
\end{footnotesize}

\paragraph{Worked example (calc, ``set A1 to 100'').}
\emph{Step 1 user:} screenshot of the empty sheet, \texttt{Instruction: set A1 to 100},
\texttt{Previous actions: None}. \emph{Step 1 assistant} (Thinking):
\texttt{<think>an MCP set-value is faster than click+type</think> Action: set A1 via MCP.
<tool\_call> \{"name": "libreoffice\_calc.set\_cell\_value", "arguments": \{"cell": "A1",
"value": 100\}\} </tool\_call>}. \emph{Step 2 user:} \texttt{Tool result:
\{"ok": true, "cell": "A1", "value": 100\}} $+$ updated screenshot $+$ \texttt{Previous
actions: Step 1: set A1 via MCP}; here the \textsc{drop} rule (Eq.~(2) of the main paper) would
replace the screenshot with $\pi$ since $c_1$ was a successful MCP call. \emph{Step 2
assistant:} \texttt{terminate/success}. Per-step outputs measure ${\sim}1500$ characters for
Thinking (with the \texttt{<think>} trace) versus ${\sim}217$ for Instruct, matching the
$6$--$7\times$ token gap reported in Section~3.1 of the main paper.

\section{Accuracy--Token Trade-off (Full Table)}
\label{sec:appendix-cost}

\begin{figure*}[tp]
\centering
\includegraphics[width=0.9\textwidth]{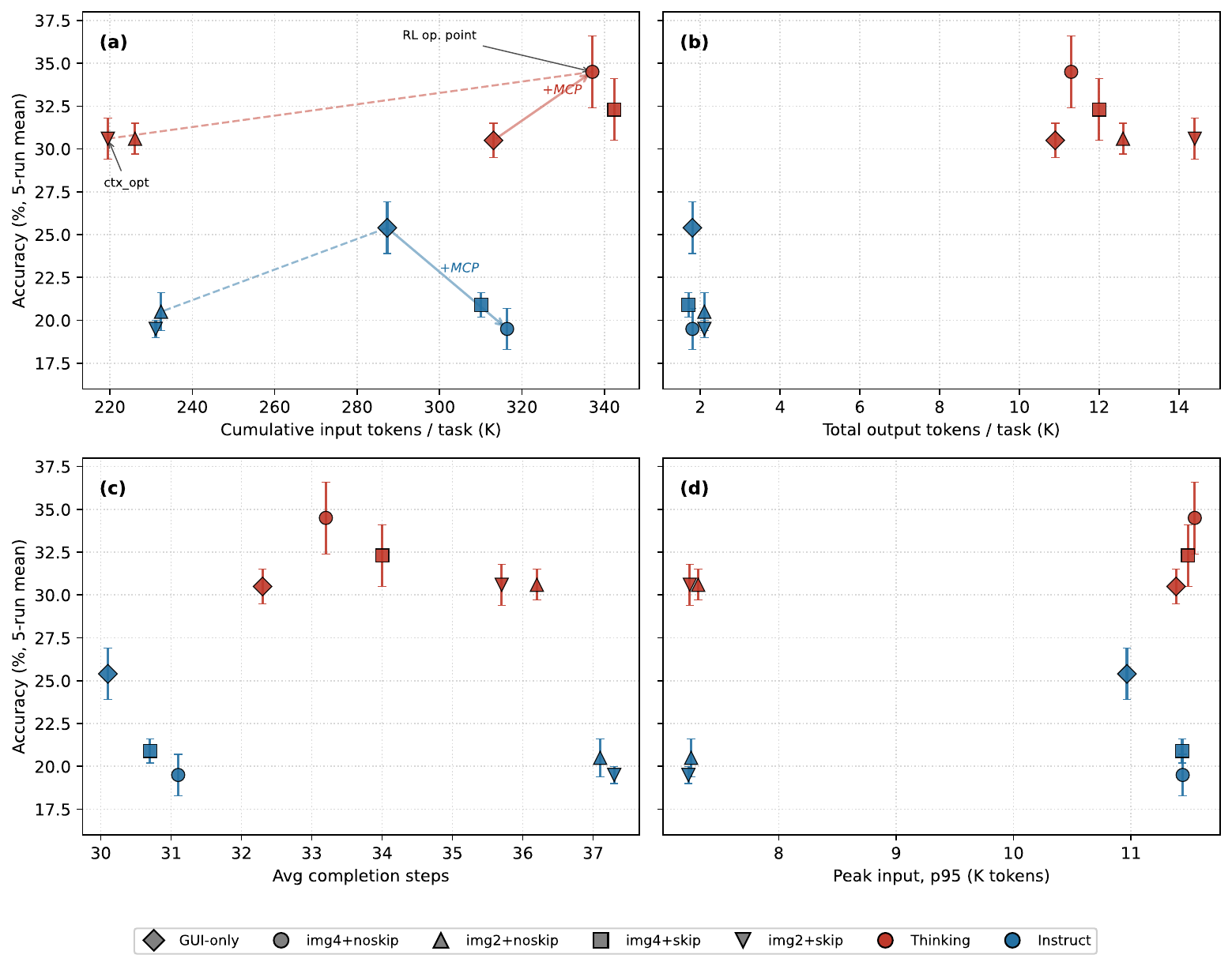}
\caption{Accuracy vs.\ cost across the five context configurations (5-run means
$\pm$std; shape = configuration, color = model). \textbf{(a)} cumulative input tokens:
the \emph{+MCP} arrows show the sign reversal; window-4 (no drop) is the RL operating point,
window-2 $+$ drop the token-efficient knee. \textbf{(b)} output tokens ($6$--$7\times$ reasoning
premium); \textbf{(c)} completion steps; \textbf{(d)} peak input (p95).}
\label{fig:pareto-supp}
\end{figure*}

Table~\ref{tab:cost} gives the exact per-configuration numbers behind the context
operating points in the token block of Table~2 of the main paper (reasoning model, mean
over 5 runs);
Figure~\ref{fig:pareto-supp} plots the full accuracy--cost breakdown for both models.

\begin{table*}[tp]
\centering
\small
\begin{tabular}{lcccc}
\toprule
Config (Thinking) & Acc & Cumul.\ input (K) & Peak input p95 & Tok / 1\%Acc (K) \\
\midrule
GUI only (window-4)        & 30.5 & 313.1 & 11385 & 10.6 \\
\textbf{window-4 (op.\ point)} & \textbf{34.5} & 337.1 & 11544 & 10.1 \\
window-2                   & 30.6 & 226.1 & 7314 & 7.8 \\
window-4 $+$ drop          & 32.3 & 342.4 & 11487 & 11.0 \\
\textbf{window-2 $+$ drop (ctx\_opt)} & \textbf{30.6} & \textbf{219.5} & \textbf{7243} & \textbf{7.7} \\
\bottomrule
\end{tabular}
\caption{Accuracy--token trade-off for the reasoning model (mean over 5 runs; peak is p95
over trajectories; Tok/1\%Acc counts input$+$output). The RL operating point is
window-4 without drop; the context-level compression is window-2 $+$ drop.}
\label{tab:cost}
\end{table*}

\section{Additional Results and Details}
\label{sec:appendix-extra}

\paragraph{Per-domain mechanism notes (Section~3.1 of the main paper).}
On writer, tasks where a tool is invoked succeed $42\%$ of the time versus $82\%$ without
a call --- a combination of difficulty self-selection (the model reaches for tools on
harder tasks) and mis-parameterized calls to writer's more complex tools --- whereas calc
and impress show almost no invoked/non-invoked success-rate difference. In multi\_apps,
tasks are often tool-unreachable because early steps run in an application whose tools do
not apply. Gimp, Thunderbird, and Chrome expose no MCP tools at all; VLC exposes 12
native tools with 16/17 tasks tool-reachable, yet neither model ever calls one.

\section{RL Training Hyperparameters}
\label{sec:appendix-hparams}

\paragraph{Hardware and infrastructure.}
All training runs use 1 node with 8 $\times$ A100 (80\,GB) GPUs under DeepSpeed ZeRO-3
in bfloat16 mixed precision. The policy model is Qwen3-VL-8B-Thinking. Environments run
in 96 parallel Docker VMs (\texttt{Ubuntu-MCP.qcow2}) mounted in \texttt{/dev/shm}.
Experiment tracking via Weights \& Biases (project \texttt{hybrid-routing-rl}).

\paragraph{Rollout.}
Each training step rolls out $G{=}8$ trajectories per task across 96 parallel environments;
with full coverage, all 74 gradient-band tasks (empirical pass rate $p\in(0.1,0.9)$ under
temperature-1.0 sampling with $k{=}8$; re-profiled as the policy improves) are rolled out
every step ($74{\times}8{=}592$ trajectories per training step). The dense-bonus run of
Section~5.2 of the main paper uses a 24-task fast-iteration subset of the band. Rollouts use
temperature $1.0$, \texttt{top\_p}$=1.0$, \texttt{max\_steps}$=50$,
\texttt{max\_tokens}$=8192$ per step, and image-history depth $k{=}4$. The MCP tool
retriever is BM25 top-18, with fallback to GUI if no tool is called. Evaluation uses greedy
decoding (temperature $0.0$) under the same observation policy as rollout; greedy probes are
repeated (3 repetitions) and judged against pre-registered criteria. For the
matched-training run of Section~5.3 of the main paper, the compression rule
(drop $+$ $k{=}2$) is active in both rollout and evaluation, with each image decision logged
and replayed in training; all other runs use $k{=}4$ without skip.

\paragraph{Reward function.}
The per-trajectory return is Eq.~(3) of the main paper: the $\pm1$ outcome term plus the
length and step-cap tie-breakers. The tool bonus $\lambda_{\text{mcp}}$ is \emph{not} part
of $R(\tau)$: it is added to the per-step advantage after group normalization (Eq.~(4) of
the main paper) and fires only for execution-successful calls of non-read-only MCP tools
whose (tool, arguments) key appears for the first time in the trajectory; it is not gated on
task outcome. Folding the bonus into $R(\tau)$ instead is a dead signal: trajectory
averaging, $z$-scoring, and the $1/T$ broadcast dilute it to ${\sim}10^{-4}$. All
coefficients are listed in Table~\ref{tab:hparams}.

\paragraph{Outcome-only sweep.}
The sweep referenced in Section~5.1 of the main paper covers learning rate
($10^{-6}$--$10^{-5}$), KL strength and anchor (base vs.\ rollout-time), normalization
scheme (group- vs.\ step-level), task density, and horizon. No configuration moves held-out
or greedy accuracy, with zero sustained fail$\to$pass flips; training-side sampled curves
separate only by KL anchor. The one exception is behavioral rather than accuracy-driven: a
rollout-time anchor combined with the dense bonus produces adoption drift that transfers to
greedy decoding with no accuracy movement (Section~5.2 of the main paper). The advantage
standard deviation is invariant at $0.068$--$0.076$ throughout.

\begin{table}[!tb]
\centering
\small
\setlength{\tabcolsep}{2pt}
\begin{tabular}{llr}
\toprule
Group & Parameter & Value \\
\midrule
\multirow{4}{*}{Reward}
  & $\lambda_{\text{len}}$ (length penalty coef) & 0.05 \\
  & $T_{\max}$ (length penalty denom.) & 50 \\
  & $\lambda_{\text{cap}}$ (step-cap penalty) & 0.20 \\
  & $\lambda_{\text{mcp}}$ (post-norm.\ step bonus) & 0.10 \\
\midrule
\multirow{6}{*}{GRPO}
  & Group size $G$ & 8 \\
  & Clip $\epsilon$ & 0.20 \\
  & KL coefficient $\beta$ & 0.02 \\
  & KL reference & rollout-time policy \\
  & KL estimator & $k_3$ \\
  & Acc.\ filter $(acc_{\text{lo}},\ acc_{\text{hi}})$ & $(0.05,\ 0.95)$ \\
\midrule
\multirow{6}{*}{Optimizer}
  & Algorithm & AdamW \\
  & Learning rate & $5{\times}10^{-6}$ \\
  & $(\beta_1,\beta_2)$ & $(0.9,\ 0.999)$ \\
  & Weight decay & 0 \\
  & LR schedule & cosine, no warmup \\
  & Max grad norm & 1.0 \\
\midrule
\multirow{5}{*}{Training}
  & Policy updates per step & 4 \\
  & Max prompt length & 16\,384 tokens \\
  & Max gen length & 8\,192 tokens \\
  & Mixed precision & bfloat16 \\
  & Random seed & 10086 \\
\bottomrule
\end{tabular}
\caption{RL hyperparameters. All reward shaping coefficients not listed are 0. The
dense-bonus run (Section~5.2 of the main paper) uses $\lambda_{\text{mcp}}{=}0.1$; the
outcome-only control and the matched-training run (Section~5.3) use
$\lambda_{\text{mcp}}{=}0$. Anchoring the KL to the base model instead of the rollout-time
policy is the ablation discussed in Section~5.1 of the main paper.}
\label{tab:hparams}
\end{table}

\end{document}